\documentclass[sigconf]{acmart}
\usepackage{pifont}
\usepackage{color}
\usepackage{multirow}
\usepackage{makecell}
\usepackage{graphicx}
\usepackage{bbding}
\usepackage{subcaption}
\usepackage{balance}
\usepackage{fancyhdr}
\makeatletter
\newcommand*\bigcdot{\mathpalette\bigcdot@{.5}}
\newcommand*\bigcdot@[2]{\mathbin{\vcenter{\hbox{\scalebox{#2}{$\m@th#1\bullet$}}}}}
\renewcommand\footnotetextcopyrightpermission[1]{}
\makeatother
\AtBeginDocument{%
  }

\setcopyright{acmcopyright}
\copyrightyear{2023}
\acmYear{2023}
\acmDOI{XXXXXXX.XXXXXXX}

\acmConference[MM '23]{Proceedings of the 31th ACM International Conference on Multimedia}{October 29--November 3, 2023}{Ottawa, Canada}
\acmPrice{15.00}
\acmISBN{978-1-4503-XXXX-X/18/06}

\acmSubmissionID{3343}

\begin{document}
	\pagestyle{plain}

\title{Exploiting Fine-Grained DCT Representations for Hiding Image-Level Messages within JPEG Images}

\author{Junxue Yang}
\affiliation{%
	\institution{College of Computer Science and Electronic Engineering, Hunan University}
	\city{Changsha}
	\country{China}
}
\email{JunxueYang@hnu.edu.cn}

\author{Xin Liao}
\affiliation{%
	\institution{College of Computer Science and Electronic Engineering, Hunan University}
	\city{Changsha}
	\country{China}
}
\email{xinliao@hnu.edu.cn}

\renewcommand{\shortauthors}{Trovato et al.}

\begin{abstract}
Unlike hiding bit-level messages, hiding image-level messages is more challenging, which requires large capacity, high imperceptibility, and high security. Although recent advances in hiding image-level messages have been remarkable, existing schemes are limited to lossless spatial images as covers and cannot be directly applied to JPEG images, the ubiquitous lossy format images in daily life. The difficulties of migration are caused by the lack of targeted design and the loss of details due to lossy decompression and re-compression. Considering that taking DCT densely on $8\times8$ image patches is the core of the JPEG compression standard, we design a novel model called \textsf{EFDR}, which can comprehensively \underline{E}xploit \underline{F}ine-grained \underline{D}CT \underline{R}epresentations and embed the secret image into quantized DCT coefficients to avoid the lossy process. Specifically, we transform the JPEG cover image and hidden secret image into fine-grained DCT representations that compact the frequency and are associated with the inter-block and intra-block correlations. Subsequently, the fine-grained DCT representations are further enhanced by a sub-band features enhancement module. Afterward, a transformer-based invertibility module is designed to fuse enhanced sub-band features. Such a design enables a fine-grained self-attention on each sub-band and captures long-range dependencies while maintaining excellent reversibility for hiding and recovery. To our best knowledge, this is the first attempt to embed a color image of equal size in a color JPEG image. Extensive experiments demonstrate the effectiveness of our \textsf{EFDR} with superior performance.
\end{abstract}




\maketitle

\section{Introduction}
Steganography is the science and art of covert communication by sheltering confidential information within publicly available objects without arousing suspicion \cite{Provos03}. This technique can date back to the 15-th century when the hidden secret information and the objects used as carriers were both physical levels. With the development of technology, modern steganography is concerned with the digital level. Many kinds of digital media such as images, videos, and audios can be used as steganographic covers, to secretly transmit digital information, among which the research on image steganography using images as carriers receives the most attention \cite{Fridrich09}. The task of image steganography is to conceal the secret messages in the cover image to generate the stego one. A well-designed steganographic scheme usually needs to satisfy three basic requirements \cite{Pevny08}: the capacity refers to the embedded payload, the imperceptibility regarding the cover and stego being indistinguishable by the naked eyes, and the security against the countermeasure, passive steganalysis.
\begin{figure}[t]
	\centering
	{\includegraphics[width=0.47\textwidth]{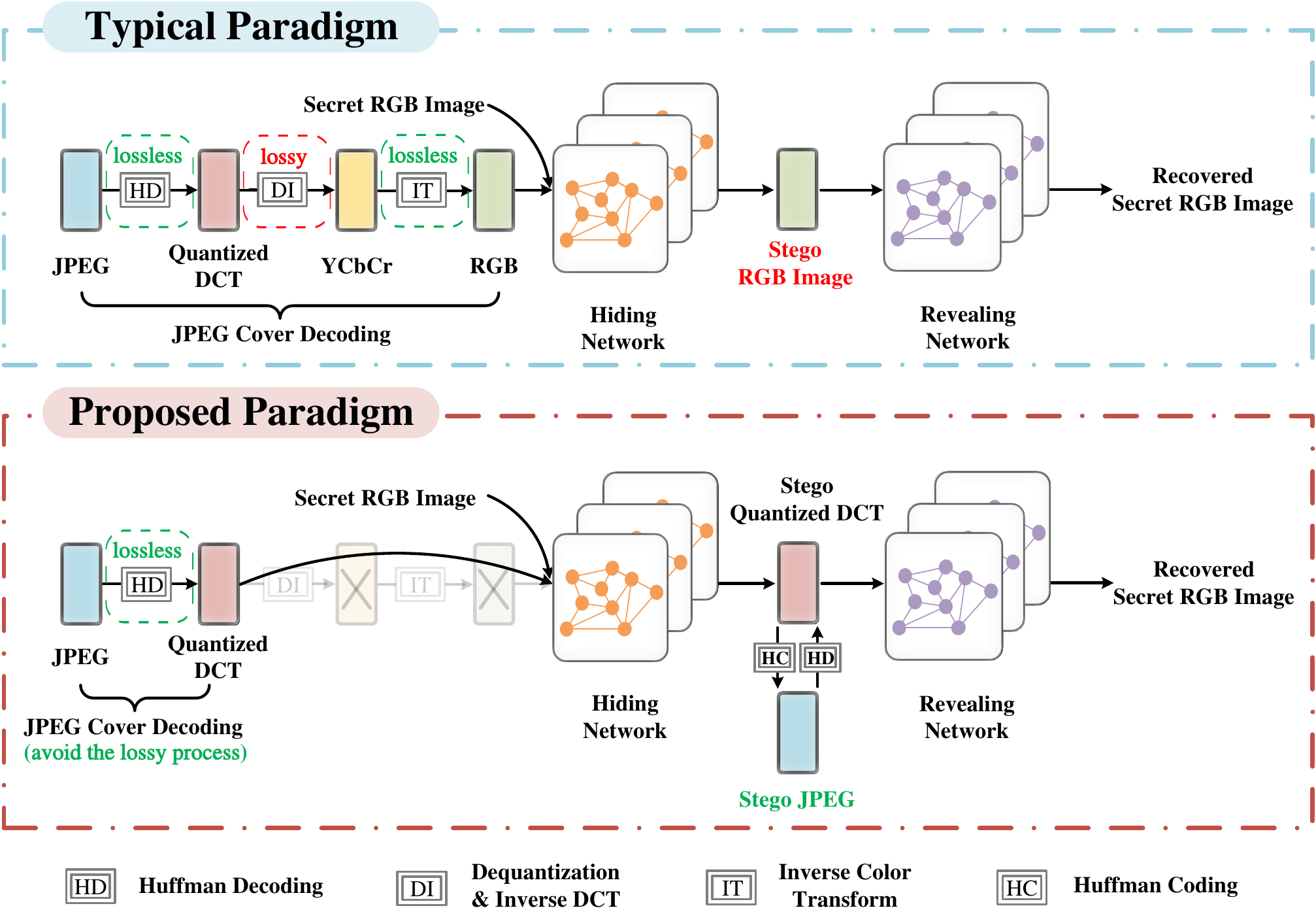}}
	\caption{The illustration of the difference between our image-level hiding method and typical image-level hiding methods when the cover is a JPEG image. Unlike typical methods (the top pipeline), we directly hide and recover secret images in the quantized DCT coefficients of the JPEG cover image and the JPEG stego image, respectively, avoiding the process that would cause information loss, and the image type of the stego obtained is also consistent with the original cover type (the bottom pipeline).}
	\label{fig:decode}
\end{figure}
On the basis of meeting these basic needs, varieties of steganographic methods \cite{Setiadi23} have been designed to pursue superior performance. Traditional content-adaptive steganographic approaches \cite{Holub12,Li15} heuristically design the distortion function considering the characteristics of the image itself, and then utilize some coding techniques, e.g., syndrome-trellis codes (STCs) \cite{Filler11}, to achieve near the minimal distortion embedding under a given payload. However, such methods typically maintain a small embedded payload, at most 1 bit per pixel \cite{Lu21}. Although several recent deep learning-based methods like SteganoGAN \cite{Zhang19} and CHAT-GAN \cite{Tan22} have shown significant improvements in hiding capacity, the magnitude of the payload is still on the bit-level, which does not comprehensively exploit the potential of deep hiding. In a seminal work \cite{Baluja17}, Baluja successfully hides a color spatial RGB image into another one of the same size, utilizing the structure of pre-processing network - hiding network - revealing network. In comparison to hiding bit-level messages, hiding image-level messages has the following natural advantages. 1) The hidden secret image does not need to be encoded perfectly due to the intrinsic relevance of the image content, which can reduce the difficulty of network learning. 2) It is also acceptable even if there are localized errors in the recovered secret image that do not significantly affect the overall appearance. 3) The amount of payload can reach up to 24 bits per pixel, as the ratio of the hidden secret image and cover image is 1:1.

The success of the pioneering work and the natural advantages of hiding image-level messages make it attract a great deal of attention. Numerous scholars are devoted to researching this field, aiming at improving hiding and revealing performance. Although existing image-level hiding works \cite{Baluja17,Baluja20,Zhang20,Lu21,Jing21} have achieved remarkable advancement, the cover type targeted by these works is only the lossless spatial RGB image. Practically, JPEG format is the most widely used and spread image format in daily life. It is a lossy format that reduces the size of the image file by compressing visually insensitive information for better storage and transmission \cite{Wallace92}, in which the information loss mainly occurs in the quantization and dequantization stages of JPEG image encoding and decoding. Even if the training and testing datasets consist of color JPEG images, typical image-level hiding methods will undergo decompression into the spatial domain before being fed into the network for learning. The obtained stego images are also in a lossless image format, ignoring the problem that the image type of the cover and stego is inconsistent. And the hidden secret image is recovered from the lossless spatial stego with the help of the revealing network (as shown in the typical paradigm of Fig. \ref{fig:decode}). Additionally, these methods lack JPEG domain knowledge in the model design, which hinders the effective extraction of relevant features. Therefore, their models may treat invisible compression artifacts as common textures, causing image quality degradation. 

Concerning these aspects, we propose that the color secret image is directly embedded into the quantized DCT coefficients of the color JPEG image and recovered from the quantized DCT coefficients of the corresponding JPEG stego image, as shown in the proposed paradigm of Fig. \ref{fig:decode}. Instead of hiding the secret image within the decompressed cover and revealing it from the re-compressed stego, our process can mitigate the information loss problem. To explore the possibility of embedding image-level information in JPEG cover images, we design a novel model called \textsf{EFDR} that aims to fully mine the fine-grained DCT representations being beneficial for hiding and recovery. It comprises three distinct designs: a fine-grained DCT representations module, a sub-band features enhancement module, and a transformer-based invertibility module. Specifically, the first is responsible for transforming the spatial secret image and quantized DCT coefficients of the JPEG cover image into frequency maps, in which each channel represents a frequency sub-band associated with inter-block correlations. And different from conventional frequency components, fine-grained frequency components can be utilized to discover more beneficial representations. The second learns the enhanced sub-band representations adaptively. Besides, to thoroughly integrate the enhanced sub-band representations of cover and secret, the third module adopts a highly reversible structure that can capture long-term dependencies, resulting in appealing outcomes. We evaluate the efficacy of the proposed model through comprehensive experiments and demonstrate its superior performance quantitatively and qualitatively over state-of-the-art (SOTA) methods while also showing its admirable generalization across different datasets.

The main contributions are summarized as follows:
\begin{itemize}
	\item To our best knowledge, we are the first to propose hiding image-level messages of equal size within a color JPEG image, the ubiquitous lossy format image in daily applications, whereas existing schemes are limited to lossless spatial images as covers. Research on hiding an image within a color JPEG image is of both academic and practical value.
	\item Unlike learning from image information that has been decompressed (for the hiding network) and then re-compressed (for the revealing network), we propose to hide and reveal the secret image in the quantized DCT coefficients of cover and stego, respectively, vastly mitigating the information loss problem.
	\item We present a powerful model for the task of hiding image-level messages into a color JPEG image with enhanced fine-grained DCT representations. And we design a sub-band features enhancement module in association with a specific DCT sub-band, and a transformer-based invertibility module combining advantages of long-range dependencies modeling of transformer and reversibility of invertible neural networks.
	\item Comprehensive experiments on three widely-used datasets show that the proposed method outperforms other SOTA methods by a large margin.
\end{itemize}

\section{Related Work}
Recently, researchers have devoted themselves to exploring the possibilities of combining deep learning with image steganography, and have made many impressive achievements. These approaches employ neural networks as either a component in the traditional algorithm (e.g., using deep learning to generate cover images more suitable for steganography \cite{Denis20,Shi18,Zhou20} and design distortion cost to measure whether the pixel is suitable for embedding \cite{Tang19,Yang20,Tang21}), or as an end-to-end solution through which secret messages can be hidden and extracted directly \cite{Hayes17,Zhang19,Wang19,Tan22,Baluja17,Baluja20,Zhang20,Lu21,Jing21}. Our work focuses on end-to-end image steganography and we summarize its recent advancement in this section. In terms of the payload magnitude, the related works are classified into embedding information at the bit-level and the image-level.

\subsection{Hiding Bit-level Messages}
Hayes et al. \cite{Hayes17} first proposed an end-to-end model to hide secret information, which defines a three-way game of hiding network, revealing network, and steganalysis network. Nonetheless, the embedding capacity of this approach is only comparable to that of traditional techniques. In addition, there are visual artifacts visible to the naked eyes. To enhance the hiding capacity, imperceptibility, and security, subsequent algorithms designed more elaborate ways to handle confidential information and introduced more advanced modules. Zhang et al. \cite{Zhang19} organized binary information into a 3-D tensor concatenated with the image tensor, which can achieve a payload of 4.4 bits per pixel, a 10-fold improvement over the previous approaches. Wang et al. \cite{Wang19} and Tan et al. \cite{Tan22} followed the information processing way and further improved security and extraction accuracy by incorporating the advanced Inception-ResNet module and channel attention mechanism, respectively. However, the embedding capacity of the aforementioned algorithms remains at the bit-level.
\begin{figure*}[t]
	\centering
	{\includegraphics[width=\textwidth]{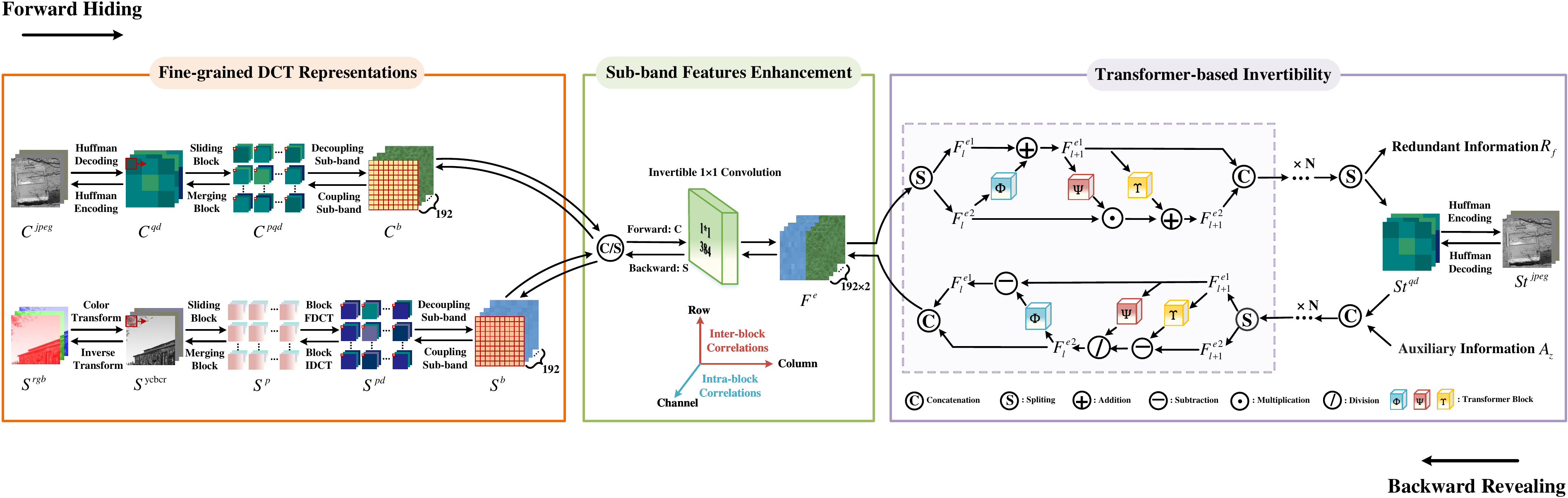}}
	\caption{Overall model architecture. In the forward hiding process, a spatial RGB secret image $S^{rgb}$ is hidden in a JPEG cover image $C^{jpeg}$ through a fine-grained DCT representations module, a sub-band features enhancement module, and a transformer-based invertibility module to generate a JPEG stego image $St^{jpeg}$, together with the redundant information $R_{f}$. In the backward revealing process, the JPEG stego image $St^{jpeg}$ and auxiliary information $A^{z}$ are fed to recover the secret image. Here, $A^{z}$ is set as an all-zero tensor. Therefore, only the stego image is actually required to extract the secret image $S^{rgb}$, ensuring blind extraction.}
	\label{fig:method}
\end{figure*}
\subsection{Hiding Image-level Messages}
Baluja \cite{Baluja17} first attempted to hide a color image within another one using deep learning. To achieve this goal, a model consisting of three modules is designed, in which the preparation network extracts fundamental features of the secret image, the hiding network fuses the extracted features into a cover image to obtain a stego image, and the revealing network restores the secret image from the stego image. Because the amount of information needed to hide is large, the risk of secret images being disclosed is correspondingly higher. The extension work \cite{Baluja20} further explored the problem that the secret image may be exposed through residuals between cover and stego. Inspired by the universal adversarial perturbations in the field of adversarial example, Zhang et al. \cite{Zhang20} designed a cover-agnostic manner to hide an image, which mitigates the risk of exposure by encoding the secret image into a universal latent representations. Lu et al. \cite{Lu21} regarded the task of hiding an image as an invertible problem in image domain transformation. They introduced the invertible network to perform embedding and extraction through its forward and backward processes, resulting in significant improvements in both visual quality and security. Jing et al. \cite{Jing21} also utilized the invertible network and implemented it in the wavelet domain. A low-frequency wavelet loss was also designed to control the embedding distribution of secret information in different frequency bands, which further promoted security. Despite that, the cover type adopted by these schemes is all the lossless spatial RGB image, without taking into account the more widely used JPEG images. Moreover, these methods lack specific design and are unsuitable when the cover is a JPEG image.

\section{Method}

In this section, we propose a novel model called \textsf{EFDR} that aims to hide image-level messages into a color JPEG image with high imperceptibility and high security. For the sake of convenience, we summarize the symbols used in this paper in Table \ref{tab:Sym}.
\begin{table}[tbh]
	\centering
	\caption{The summary of symbols used in this paper.}
	\label{tab:Sym}
	\belowrulesep=0pt
	\aboverulesep=0pt
	\renewcommand{\tabcolsep}{0.3mm} 
	\renewcommand{\arraystretch}{1.5}
	\resizebox{0.48\textwidth}{!}
	{\begin{tabular}{c|c}
		\toprule
		Symbols & Descriptions \\ \midrule
		$C^{jpeg}$ & JPEG cover image: the image to hide secret information \\
		$C^{qd}$ & quantized DCT coefficients of JPEG cover \\
		$C^{pqd}$ & a set of $8\times8$ quantized DCT coefficients patches of JPEG cover \\
		$C^b$ & decoupling frequency maps of JPEG cover \\ \cline{1-2}
		$S^{rgb}$ & spatial RGB secret image: the image to be hidden \\ 
		$S^{ycbcr}$ & the YCbCr transformation of secret \\ 
		$S^{p}$ & a set of $8\times8$ YCbCr transformation patches of secret \\ 
		$S^{pd}$ & a set of $8\times8$ DCT coefficients patches of secret \\ 
		$S^{b}$ & decoupling frequency maps of secret \\  \cline{1-2}
		$St^{jpeg}$ & JPEG stego image: the image containing secret information \\
		$St^{qd}$ & quantized DCT coefficients of JPEG stego \\
		$F^{e}$ & the enhanced sub-band feature maps \\
		$R_{f}$ & redundant information: the lost information in hiding process \\
		$A^{z}$ &  auxiliary information: the information to help reveal secret \\
		\bottomrule
	\end{tabular}
}
\end{table}
\subsection{Overview}
Fig. \ref{fig:method} shows the framework of the proposed \textsf{EFDR}, containing three key components: a fine-grained DCT representations module, a sub-band features enhancement module, and a transformer-based invertibility module. In the forward hiding process, a pair of JPEG cover image $C^{jpeg}$ and spatial RGB secret image $S^{rgb}$ are as the input. They are converted into fine-grained frequency maps, in which each channel denotes a frequency sub-band. Then, after sub-band features enhancement and a series of transformer-based invertible sub-modules, the quantized DCT coefficients of JPEG stego $St^{qd}$ can be obtained, together with the lost redundant information $R_{f}$. $St^{qd}$ can be transformed into JPEG image $St^{jpeg}$ by Huffman coding. Conversely, in the backward revealing process, $St^{jpeg}$ and auxiliary information $A^{z}$ go through the inverse process of the above three modules to recover $S^{rgb}$. The hiding and revealing processes can be expressed as
\begin{gather}	
	(St^{jpeg},R_{f})=\textbf{T}(\textbf{E}(\textbf{F}(C^{jpeg},S^{rgb}))) \nonumber \\
	(C^{jpeg},S^{rgb})=\textbf{F}^{-1}(\textbf{E}^{-1}(\textbf{T}^{-1}(St^{jpeg},A^{z})))
	\label{eq1}
\end{gather}
In Eq. (\ref{eq1}), $\textbf{F}(\bigcdot)$, $\textbf{E}(\bigcdot)$, $\textbf{T}(\bigcdot)$ and $\textbf{F}^{-1}(\bigcdot)$, $\textbf{E}^{-1}(\bigcdot)$, $\textbf{T}^{-1}(\bigcdot)$ represent the three modules and their inverse processes respectively. The modules details and loss functions will be further provided in the following.

\subsection{Fine-grained DCT Representations}
As shown in the leftmost of Fig. \ref{fig:method}, we convert the JPEG cover image and secret RGB image into fine-grained frequency components that facilitate hiding and revealing. The processing of JPEG cover image $C^{jpeg}\in{\mathbb{R}^{(3,H,W)}}$ and secret RGB image $S^{rgb}\in{\mathbb{R}^{(3,H,W)}}$ is slightly different, where the superscript $(3,H,W)$ denote the shape with $(channel,height,width)$. Concretely, for $C^{jpeg}$, the quantized DCT coefficients $C^{qd}\in{\mathbb{R}^{(3,H,W)}}$ is firstly extracted by Huffman decoding. After that, we can acquire a set of quantized DCT coefficients blocks $C^{pqd}\in{\mathbb{R}^{(3\times{\frac{H}{8}}\times{\frac{W}{8}},8,8)}}$ through a fixed size $8\times8$ sliding window. Each value in $C^{pqd}$ corresponds to the intensity of 192 frequency bands of 3 color channels. To make the frequency components more compact, we flatten and reshape $C^{pqd}$ to form fully decoupling frequency maps $C^b\in{\mathbb{R}^{(192,\frac{H}{8},\frac{W}{8})}}$, where the values in each channel represent the intensity of a frequency sub-band. In this way, the original input is reconstructed to fine-grained frequency-aware information that contains both the intra-block correlations in the channel dimension and the inter-block correlations in the space dimension. For the processing of the $S^{rgb}$, there are two more steps, namely color conversion to YCbCr space to obtain $S^{ycbcr}\in{\mathbb{R}^{(3,H,W)}}$ and dense DCT transform on $8\times8$ YCbCr transformation patches $S^{p}\in{\mathbb{R}^{(3\times{\frac{H}{8}}\times{\frac{W}{8}},8,8)}}$ to attain DCT coefficients patches, denoted by $S^{pd}\in{\mathbb{R}^{(3\times{\frac{H}{8}}\times{\frac{W}{8}},8,8)}}$. In the backward revealing process, just perform the inverse operations of the previous procedures.

\subsection{Sub-band Features Enhancement}
The structure of this module is shown in the middle of Fig. \ref{fig:method}, we first concatenate the decoupling frequency maps $C^b$ and $S^b$ in the channel dimension. Subsequently, an invertible $1\times1$ convolution layer \cite{Kingma18} is applied to their concatenation to adaptively extract the most informative frequency features. The invertible $1\times1$ convolution is a kind of convolution commonly used for flow-based generative models. It performs a linear transformation on the input feature maps by multiplying them with a random rotation weight matrix with the nonzero determinant. And its inverse operation is to multiply the output feature maps with the inverse of the weight matrix to restore the input feature maps. Here, the number of convolution kernels is set to 384, which is equal to the sum of the DCT frequency modes of the color JPEG image and the secret image. Such a design can make each element in the enhanced feature maps (denoted by $F^e\in{\mathbb{R}^{(384,\frac{H}{8},\frac{W}{8})}}$) associated with a specific DCT sub-band.
\begin{figure*}[t]
	\centering
	{\includegraphics[width=\textwidth]{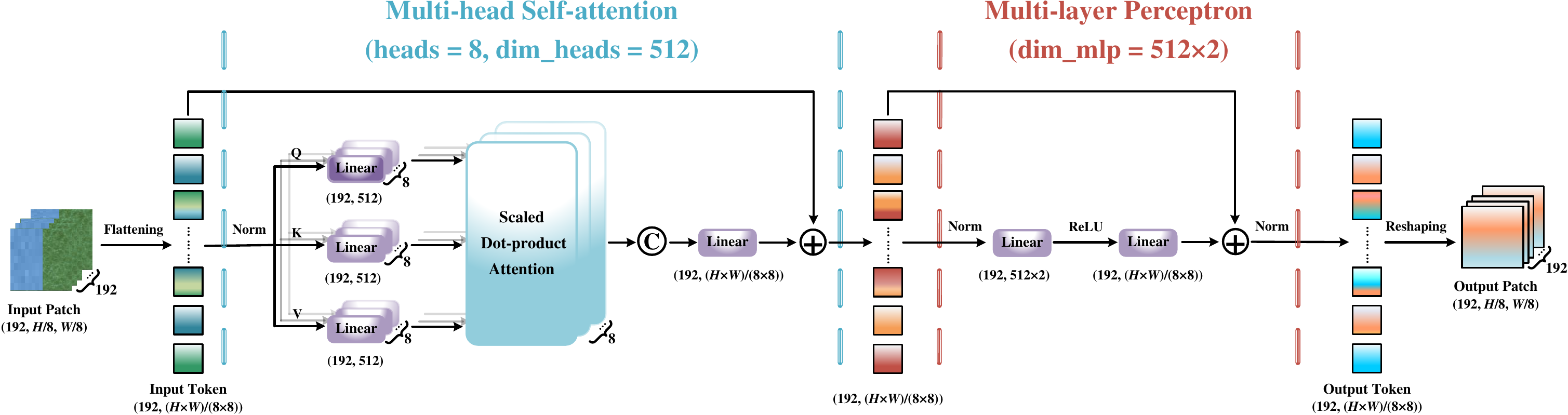}}
	\caption{The detailed structure of our $\Phi(\protect\bigcdot)$, $\Psi(\protect\bigcdot)$, and $\Upsilon(\protect\bigcdot)$, i.e., a 1-layer Pre-LN transformer block, which consists of a multi-head self-attention and a multi-layer perceptron. Concretely, the dimension of input and output are the same. We employ $heads=8$ parallel attention layers, as well as the dimension $dim\_heads=512$ of the inner layer. And in the multi-layer perceptron, the dimension $dim\_mlp=512\times2$ of inner layer.}
	\label{fig:trans}
\end{figure*}
\subsection{Transformer-based Invertibility}
Our transformer-based invertibility module, shown in the rightmost of Fig. \ref{fig:method}, is derived from invertible neural networks \cite{Dinh17,Xiao20} and the vision transformer \cite{Dosovitskiy21}, both of which are good at image-related tasks. Currently, many image-related tasks have employed invertible neural networks, such as image re-scaling \cite{Xiao20}, image decolorization \cite{Zhao21}, and image translation \cite{Ouderaa19}. The vision transformer is also widely applied to image-related fields, including image classification \cite{Liu21}, semantic segmentation \cite{Zheng21}, and object detection \cite{Carion20}. An invertible sub-module adopts the bijective affine coupling layer with shared parameters to perform forward and backward operations so that the details can be preserved as much as possible to achieve the reversibility of input and output. To further enhance representation and global dependency modeling, the transformer is introduced as the backbone. To be more specific, for the $l$-th invertible sub-module, $1\leq{l}\leq{N}$, we adopt an additive transform for branch 1 ($F^{e1}_l\in{\mathbb{R}^{(192,\frac{H}{8},\frac{W}{8})}}$) and an enhanced affine transform for branch 2 ($F^{e2}_l\in{\mathbb{R}^{(192,\frac{H}{8},\frac{W}{8})}}$). Thus, the forward process of the bijective affine coupling layer can be formulated as
\begin{gather}	
	F^{e1}_{l+1}=F^{e1}_{l}+\Phi(F^{e2}_{l}) \nonumber \\
	F^{e2}_{l+1}=F^{e2}_{l}\odot{\textbf{exp}(\Psi(F^{e1}_{l+1}))}+\Upsilon(F^{e1}_{l+1})
	\label{eq2}
\end{gather}
In Eq. (\ref{eq2}), $\Phi(\bigcdot)$, $\Psi(\bigcdot)$, and $\Upsilon(\bigcdot)$ denote the transformer block. Here, we adopt a 1-layer Pre-LN transformer, as shown in
Fig. \ref{fig:trans}. The Pre-LN transformer, which puts the layer normalization before the multi-head self-attention and multi-layer perceptron, compared with the Post-LN transformer, has a more stable gradient distribution, and the gradient norm is almost constant between layers, which is more conducive to optimization by the optimizer \cite{Xiong20}. $\odot$ and $\textbf{exp}(\bigcdot)$ are the Hadamard product and Exponential function, respectively. Correspondingly, the inverse operation is
\begin{gather}	
	F^{e2}_{l}=(F^{e2}_{l+1}-\Upsilon(F^{e1}_{l+1}))\odot\textbf{exp}(-\Psi(F^{e1}_{l+1})) \nonumber \\
	F^{e1}_{l}=F^{e1}_{l+1}-\Phi(F^{e2}_{l})
	\label{eq3}
\end{gather}
Note that $\textbf{exp}(\bigcdot)$ is omitted in the figure.

\subsection{Loss Functions}
The total loss $\mathcal{L}_{total}$ consists of two parts: the hiding loss $\mathcal{L}_{hi}$ to ensure imperceptibility and security, and the revealing loss $\mathcal{L}_{re}$ to ensure the recovery quality. $\mathcal{L}_{total}$ is equal to the sum of the two losses.

\textbf{Hiding loss.} The forward hiding process aims to embed $S^{rgb}$ into $C^{jpeg}$ to generate $St^{jpeg}$. To achieve imperceptibility and security, $St^{jpeg}$ is required to be close to $C^{jpeg}$. Thus, $\mathcal{L}_{hi}$ is defined as follows
\begin{gather}	
	\mathcal{L}_{hi}=MSE(St^{jpeg},C^{jpeg})
	\label{eq4}
\end{gather}

\textbf{Revealing loss.} The backward revealing process aims to losslessly extract $S^{rgb}$ from $St^{jpeg}$. Toward this goal, we define the $\mathcal{L}_{re}$ as follows
\begin{gather}	
	\mathcal{L}_{re}=MSE(S^{rgb}_{ex},S^{rgb})
	\label{eq5}
\end{gather}
where $S^{rgb}_{ex}$ represents the actual extracted secret image. It should be noted that we do not set any restrictions on the forward redundant information $R_{f}$ and the recovered cover JPEG image, because these two variables are less important in the whole hiding and revealing tasks.

\section{Experiments}
\subsection{Experimental Settings}
\textbf{Datasets.} We conduct experiments on three image datasets: COCO \cite{Lin14} and ImageNet \cite{Russakovsky15}, and a standard steganography research dataset BOSSBase \cite{Bas11}. 10000 and 4000 images are randomly selected from COCO as the training set and test set, respectively. ImageNet and BOSSBase only serve as test sets, and the number of images is also 4000. All images are center-cropped to the size of $128\times{128}$ and re-compressed into JPEG format with two commonly used quality factors (QFs) 75 and 95. Although some images in these datasets may suffer from double/multiple JPEG compression, their use does not jeopardize the feasibility of the proposed scheme \cite{Tang19}. Double/multiple JPEG compression images are more common in practice.

\textbf{Benchmarks.} To verify the effectiveness of our method, we compare it with SOTA image-level hiding works that hide a color image into another one of equal size, including Baluja \cite{Baluja20}, UDH \cite{Zhang20}, ISN \cite{Lu21}, and HiNet \cite{Jing21}. For a fair comparison, we retrain and evaluate the models using the same datasets as ours, with the parameters following the default settings in these references.

\textbf{Evaluation metrics.} Four metrics are adopted to measure the image quality of cover/stego and secret/recovery pairs, such as peak signal-to-noise ratio (PSNR) and structural similarity (SSIM), two commonly used objective evaluation indicators \cite{Zhou04}, average pixel discrepancy (APD) calculated by $L1$ norm, and learned perceptual image patch similarity (LPIPS), which is more consistent with human visual perception \cite{Zhang18}. PSNR /SSIM with larger values and APD/LPIPS with smaller values indicate higher image quality. In addition, we employ two representative JPEG steganalyzers to evaluate the security performance of our method, including handcrafted features-based DCTR \cite{Holub15} and deep learning-based SRNet \cite{Boroumand19}. 

\textbf{Implementation details.}
Our \textsf{EFDR} is implemented with PyTorch, and the NVIDIA GeForce RTX 2080 Ti GPU is used for acceleration. We adopt the Adam optimizer with $betas=(0.5, 0.999)$, $eps=1e^{-6}$, and $weight\_decay=5e^{-4}$. The initial learning rate is $5e^{-4}$, which is dynamically adjusted by the ReduceLROnPlateau learning strategy with the factor of 0.5. The mini-batch size is 4, containing two randomly selected cover/secret pairs. And the number of total training epoch is 200. Both $QF=75$ and $QF=95$, our model contains 12 transformer-based invertible sub-modules, and each of them uses three 1-layer Pre-LN transformer blocks.
\begin{table*}[tbh]
	\centering
	\belowrulesep=0pt
	\aboverulesep=0pt
	\caption{Quantitative quality comparisons on different datasets with QFs of 75 and 95. Value1/value2 represent the assessment value calculated by the cover/stego and secret/recovery image pairs, respectively, under the corresponding metrics. The best and second-best results separately are marked in bold and underline.}
	\label{tab:QC}
	\renewcommand{\tabcolsep}{1.3mm} 
	\renewcommand{\arraystretch}{2.2}
	\resizebox{\textwidth}{!}
	{
		\begin{tabular}{cc|cccc|cccc|cccc}
			\toprule
			\multirow{2}{*}{QFs} &
			\multirow{2}{*}{Methods} &
			\multicolumn{4}{c}{COCO}\vline &
			\multicolumn{4}{c}{ImageNet}\vline &
			\multicolumn{4}{c}{BOSSBase} \\
			\cmidrule(r){3-14}
			\multicolumn{1}{c}{} &
			\multicolumn{1}{c}{}\vline
			& PSNR(dB)$\uparrow$ & SSIM$\uparrow$ & APD$\downarrow$ & LPIPS$\downarrow$ & PSNR(dB)$\uparrow$ & SSIM$\uparrow$ & APD$\downarrow$ & LPIPS$\downarrow$	&  PSNR(dB)$\uparrow$ & SSIM$\uparrow$ & APD$\downarrow$ & LPIPS$\downarrow$ \\ \midrule
			\multirow{5}{*}{75} & Baluja \cite{Baluja20} & 24.24/21.55 & \underline{.847}/.754 & 11.54/18.86 & .188/.393 & 24.68/21.55 & \underline{.851}/.746 & 11.75/19.30 & .182/.389 & \underline{26.82}/23.38 & \underline{.842}/.820 & 8.82/16.49 & \underline{.217}/.409 \\
			\multirow{5}{*}{} & UDH \cite{Zhang20} & 23.94/21.89 & .820/.748 & 8.62/18.26 & \underline{.173}/.301 & 24.69/21.77 & .825/.737 & \underline{8.48}/18.98 & \underline{.168}/.306 & 24.81/23.54 & .791/.805 & 8.62/16.30 & .252/.302 \\
			\multirow{5}{*}{} & ISN \cite{Lu21} & \underline{24.43}/\underline{24.47} & .834/\underline{.858} & \underline{8.54}/\underline{7.51} & .180/\underline{.169} & \underline{25.16}/\underline{24.88} & .836/\underline{.859} & 8.63/\underline{7.62} & .177/\underline{.170} & 26.03/\underline{29.30} & .821/\underline{.909} & \underline{7.86}/\underline{5.94} & .234/\underline{.120} \\
			\multirow{5}{*}{} & HiNet \cite{Jing21} & 16.90/20.17 & .489/.662 & 21.50/16.37 & .444/.407 & 17.43/20.42 & .504/.659 & 20.87/16.67 & .433/.408 & 17.26/23.21 & .446/.751 & 21.14/12.55 & .561/.389 \\
			\multirow{5}{*}{} & \textsf{EFDR} & \textbf{32.16}/\textbf{35.30} & \textbf{.928}/\textbf{.934} & \textbf{4.72}/\textbf{3.70} & \textbf{.024}/\textbf{.017} & \textbf{31.92}/\textbf{35.38} & \textbf{.923}/\textbf{.935} & \textbf{4.94}/\textbf{3.78} & \textbf{.027}/\textbf{.018} & \textbf{35.90}/\textbf{37.17} & \textbf{.944}/\textbf{.950} & \textbf{3.12}/\textbf{2.96} & \textbf{.028}/\textbf{.014} \\ \hline \hline
			\multirow{5}{*}{95} & Baluja \cite{Baluja20} & \underline{24.44}/21.58 & \underline{.859}/.754 & 10.92/18.81 & .183/.393 & 24.88/21.58 & \underline{.860}/.746 & 11.23/19.23 & .178/.389 & \underline{27.03}/23.40 & \underline{.840}/.821 & 8.12/16.54 & \underline{.196}/.408 \\
			\multirow{5}{*}{} & UDH \cite{Zhang20} & 24.25/21.87 & .818/.738 & \underline{7.38}/18.30 & \underline{.159}/.307 & 24.96/21.61 & .825/.719 & \underline{7.24}/19.29 & \underline{.151}/.319 & 25.12/23.18 &  .797/.764 & \underline{7.78}/16.65 & .225/.325 \\
			\multirow{5}{*}{} & ISN \cite{Lu21} & 24.42/\underline{23.38} & .843/\underline{.832} & 7.60/\underline{9.01} & .178/\underline{.201} & \underline{25.08}/\underline{23.73} & .843/\underline{.831} & 7.77/\underline{9.46} & .176/\underline{.202} & 25.42/\underline{27.43} & .812/\underline{.879} & 7.86/\underline{7.39} & .230/\underline{.161} \\
			\multirow{5}{*}{} & HiNet \cite{Jing21} & 17.59/20.03 & .543/.611 & 17.85/16.82 & .380/.448 & 18.14/20.27 & .556/.608 & 17.29/17.04 & .373/.450 & 17.99/22.52 & .509/.679 & 17.65/13.52 & .473/.424 \\
			\multirow{5}{*}{} & \textsf{EFDR} & \textbf{34.25}/\textbf{32.08} & \textbf{.925}/\textbf{.887} & \textbf{3.89}/\textbf{5.17} & \textbf{.013}/\textbf{.048} & \textbf{34.03}/\textbf{32.16} & \textbf{.920}/\textbf{.891} & \textbf{4.08}/\textbf{5.21} & \textbf{.013}/\textbf{.046} & \textbf{35.71}/\textbf{33.66} & \textbf{.909}/\textbf{.906} & \textbf{3.30}/\textbf{4.30} & \textbf{.018}/\textbf{.053} \\
			\bottomrule
		\end{tabular}
	}
\end{table*}
\subsection{Comparisons with SOTA Methods}
\textbf{Quantitative quality evaluation.} Table \ref{tab:QC} reports the numerical comparisons of our \textsf{EFDR} with Baluja \cite{Baluja20}, UDH \cite{Zhang20}, ISN \cite{Lu21}, and HiNet \cite{Jing21} in terms of the four image quality metrics. It can be observed that the proposed \textsf{EFDR} significantly outperforms other approaches in terms of both hiding and revealing performance. For $QF=75$, compared with the second-best results, our \textsf{EFDR} can achieve a PSNR improvement of 7.73/10.83 dB, 6.76/10.50 dB, and 9.08/7.87 dB on COCO, ImageNet, and BOSSBase datasets, respectively. Regarding $QF=95$, we supply 9.81/8.70 dB, 8.95/8.43 dB, and 8.68/6.23 dB PSNR advancement. Similar performance betterment can be seen in SSIM, APD, and LPIPS. It is noteworthy that although our model only uses the COCO dataset for training, it still provides excellent results on the other two datasets, which indicates that our model has a favorable generalization ability.
\begin{figure*}[t]
	\centering
	{\includegraphics[width=\textwidth]{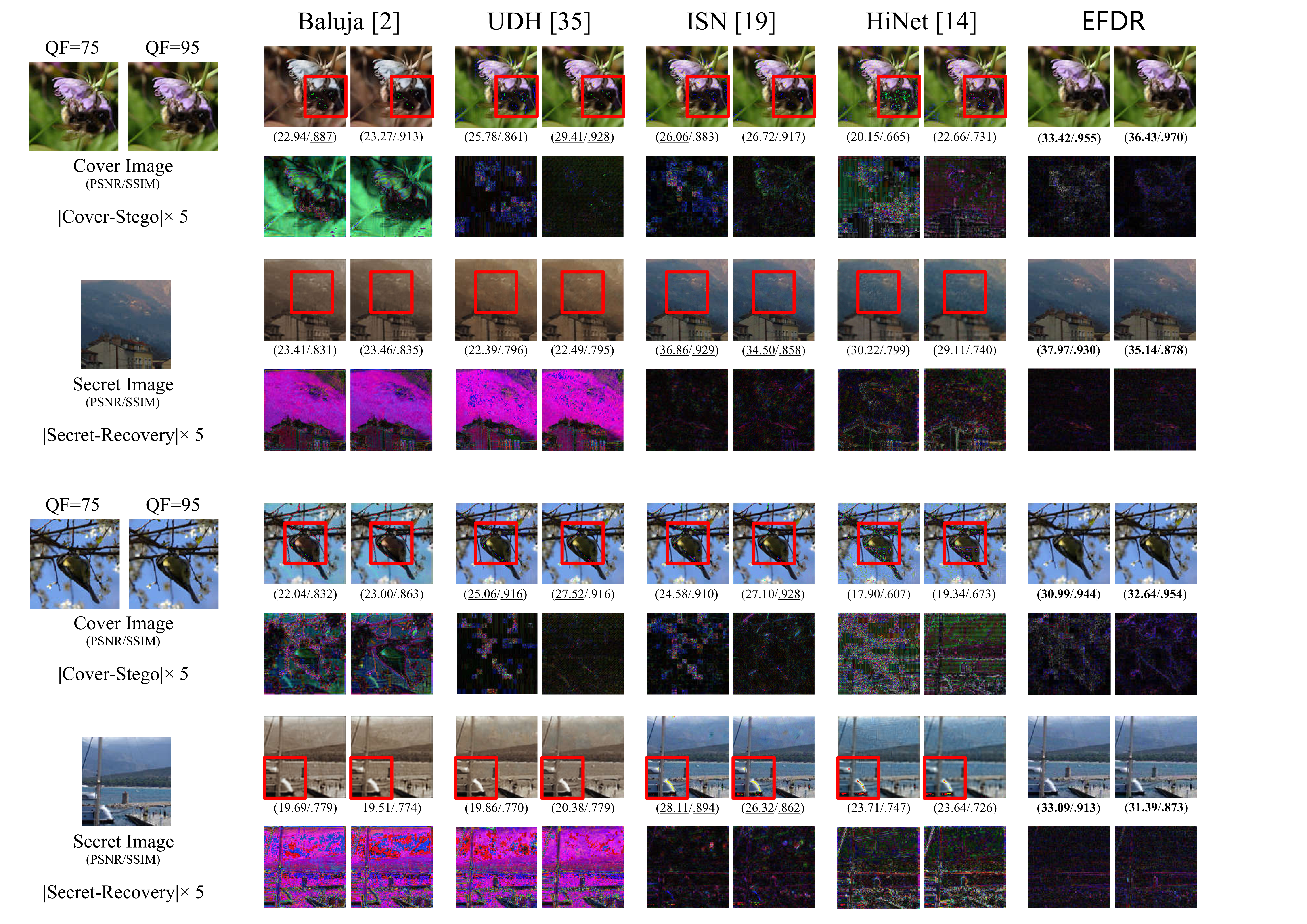}}
	\caption{Visualization comparisons of our \textsf{EFDR} with Baluja \cite{Baluja20}, UDH \cite{Zhang20}, ISN \cite{Lu21}, and HiNet \cite{Jing21}. To illustrate the differences between the original and generated images, the pixel-wise residuals are magnified by 5 times. The red boxes mark the color distortion, blurring, noise, and other artifacts that are visible to the naked eyes in the stegos and the recovered secret images generated by the comparison methods.}
	\label{fig:qv}
\end{figure*}

\begin{figure}[htb]
	\centering
	{\includegraphics[width=0.4\textwidth]{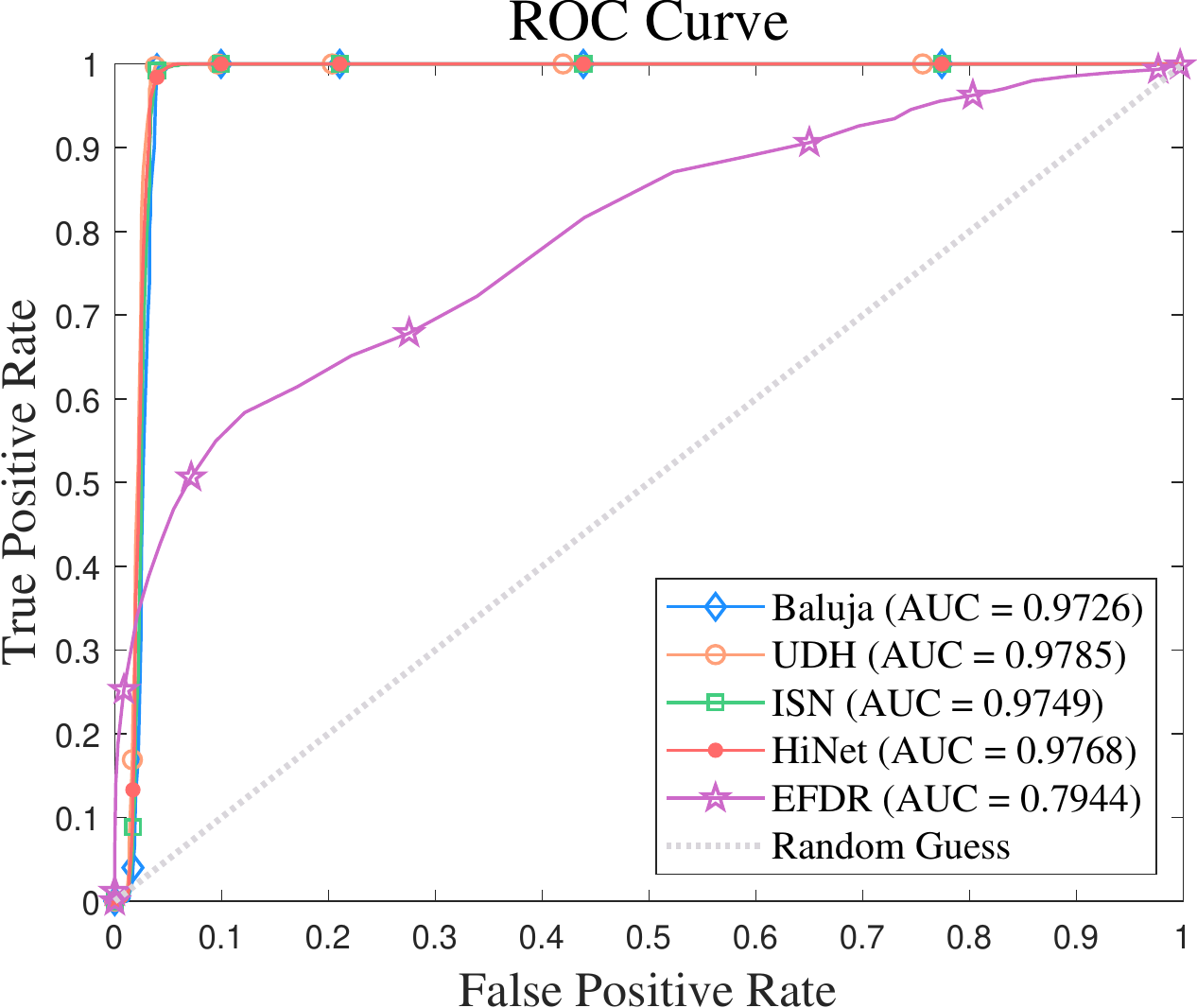}}
	\caption{Comparisons of anti-steganalyzer DCTR when QF is 75.}
	\label{fig:ROC75}
\end{figure}

\begin{figure}[htb]
	\centering
	{\includegraphics[width=0.4\textwidth]{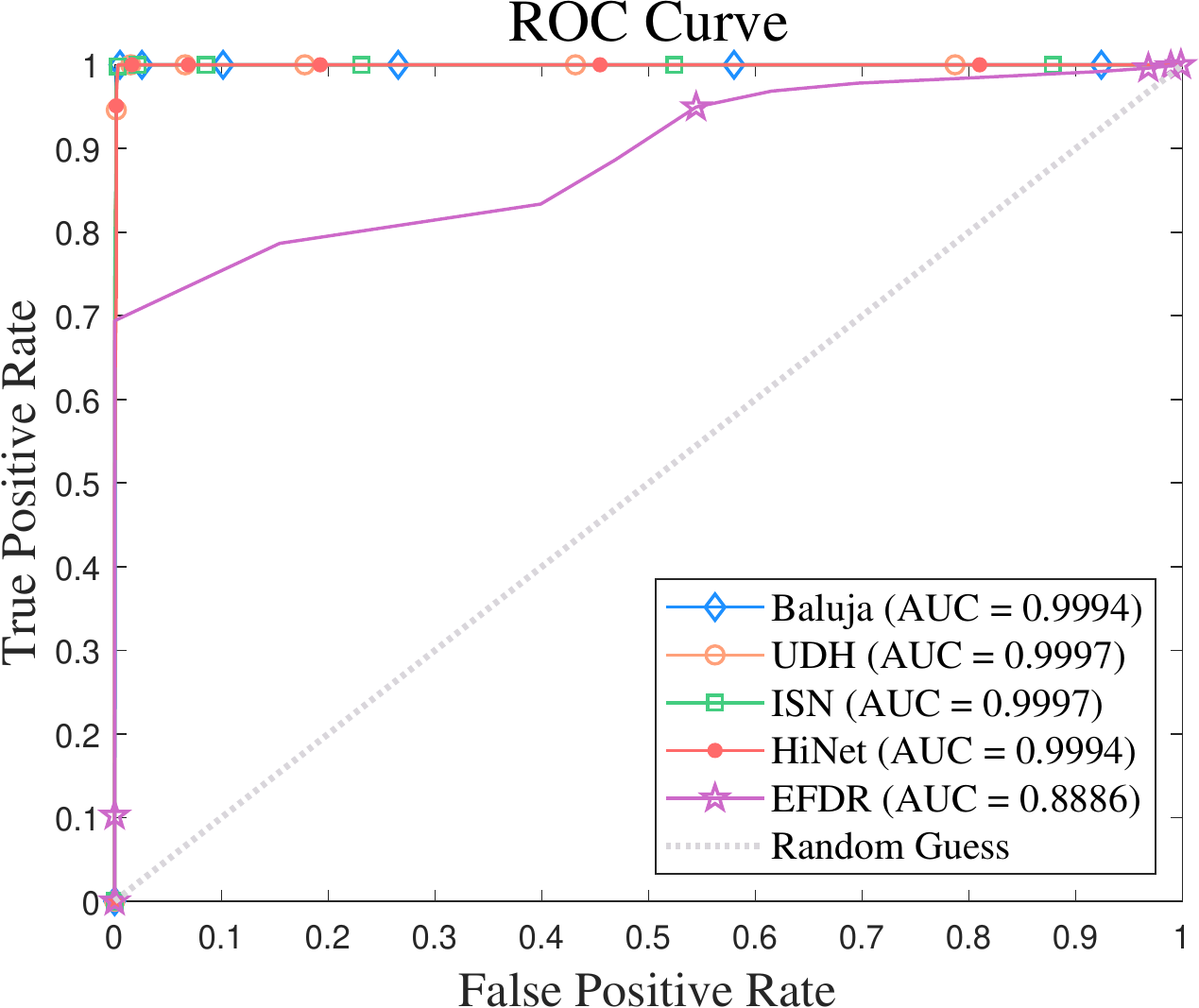}}
	\caption{Comparisons of anti-steganalyzer DCTR when QF is 95.}
	\label{fig:ROC95}
\end{figure}

\begin{table}[htb]
	\centering
	\caption{Comparisons of anti-steganalyzer SRNet.}
	\label{table:SRNet}
	\renewcommand{\tabcolsep}{4mm} 
	\renewcommand{\arraystretch}{1.2}
	\begin{tabular}{ccc}
		\toprule
		\multirow{2}{*}{Methods} & \multicolumn{2}{c}{Accuracy(\%)$\downarrow$ $\pm$ std} \\
		\cmidrule(r){2-3}
		\multicolumn{1}{c}{} & QF=75 & QF=95 \\ \midrule
		Baluja \cite{Baluja20} & 90.93 $\pm$ .196 & 97.80 $\pm$ .105 \\
		UDH \cite{Zhang20} & 98.73 $\pm$ .079 & 99.93 $\pm$ .019 \\
		ISN \cite{Lu21} & 91.43 $\pm$ .189 & 97.80 $\pm$ .104 \\
		HiNet \cite{Jing21} & 99.50 $\pm$ .050 & 99.95 $\pm$ .016 \\
		\textsf{EFDR} & \textbf{80.32 $\pm$ .269} & \textbf{82.05 $\pm$ .255} \\
		\bottomrule
	\end{tabular}
\end{table}

\begin{table}[t]
	\centering
	\belowrulesep=0pt
	\aboverulesep=0pt
	\caption{Ablation Study on different numbers of sub-module selection in transformer-based invertibility module.}
	\label{table:SS}
	\renewcommand{\tabcolsep}{0.05mm} 
	\renewcommand{\arraystretch}{1.6}
	\begin{tabular}{cccccc}
		\toprule
		& Nums. & PSNR(dB)$\uparrow$ & SSIM$\uparrow$ & APD$\downarrow$ & LPIPS$\downarrow$ \\ \midrule
		\multirow{4}{*}{QF=75} & 16 & 9.70/15.83 & .261/.497 & 49.55/21.67 & .366/.433 \\
		\multirow{4}{*}{} & 12 & \textbf{31.92}/\textbf{35.38} & \textbf{.923}/\textbf{.935} & \textbf{4.94}/\textbf{3.78} & \textbf{.027}/\textbf{.018} \\
		\multirow{4}{*}{} & 8 & 28.98/31.27 & .886/.912 & 6.06/4.67 & .053/.031 \\
		\multirow{4}{*}{} & 4 & 27.52/24.26 & .772/.643 & 8.01/12.64 & .146/.289 \\ \cmidrule(r){1-6}
		\multirow{4}{*}{QF=95} & 16 & 32.01/27.22 & .793/.774 & 6.98/8.78 & .031/.077 \\
		\multirow{4}{*}{} & 12 & \textbf{34.03}/\textbf{32.16} & \textbf{.920}/\textbf{.891} & \textbf{4.08}/\textbf{5.21} & \textbf{.013}/\textbf{.046} \\
		\multirow{4}{*}{} & 8 & 10.49/13.45 & .134/.306 & 68.70/37.19 & .625/.504 \\
		\multirow{4}{*}{} & 4 & 30.18/29.95 & .843/.849 & 6.37/6.53 & .036/.047 \\
		\bottomrule
	\end{tabular}
\end{table}
\textbf{Qualitative quality evaluation.} Visualization comparisons between our \textsf{EFDR} and the other four methods are shown in Fig. \ref{fig:qv}. Two examples shown are from ImageNet and BOSSBase datasets. We can observe that both the generated stego image and the recovered secret image are visually pleasing, with smaller visual residuals, which indicates that our model has nice hiding and revealing performance. In contrast, the comparison methods have difficulties in effectively extracting relevant features as they learn from image information that has been decompressed (for the hiding network) and then re-compressed (for the revealing network), and lack the utilization of JPEG image domain knowledge in their model design. Instead, the information differences caused by decompression and re-compression may be amplified, resulting in visible color distortion, blocking, blurring, and banding artifacts in the generated stego and recovered secret image.

\textbf{JPEG steganalysis.} Steganalysis, another essential metric, evaluates the statistical security for hiding methods. It measures the likelihood of distinguishing stego images from cover images by revealing alterations in statistical features caused by embedding. Currently, mainstream steganalysis methods can be divided into two categories: traditional handcrafted features-based methods and deep learning-based methods. In this subsection, we select two representative JPEG image steganalyzers to evaluate the anti-steganalysis ability of our \textsf{EFDR} and the comparison methods, including handcrafted features-based DCTR \cite{Holub15} and deep learning-based SRNet \cite{Boroumand19}. Specifically, we select 2000 pairs of cover images from BOSSBase dataset and the corresponding stego images generated by different methods for steganalysis security evaluation. Fig. \ref{fig:ROC75}-\ref{fig:ROC95} and Table \ref{table:SRNet} show the ROC curve comparisons and the detection accuracy comparisons of anti-steganalyzer DCTR and SRNet. The area under the ROC curve (AUC) and detection accuracy with smaller values indicate higher security. We can see that the stego images generated by our model always have higher anti-steganalysis ability.
\begin{table*}[tbh]
	\centering
	\belowrulesep=0pt
	\aboverulesep=0pt
	\caption{Ablation Study for different modules. Freq means the common DCT transform. $\textbf{F}(\protect\bigcdot)$ and $\textbf{E}(\protect\bigcdot)$ represent the fine-grained DCT representations module and sub-band features enhancement module, respectively.}
	\label{table:AC}
	\renewcommand{\tabcolsep}{5mm} 
	\renewcommand{\arraystretch}{1.4}
	\begin{tabular}{c|ccc|cccc}
		\toprule
		& Freq & $\textbf{F}(\bigcdot)$ & $\textbf{E}(\bigcdot)$ & PSNR(dB)$\uparrow$ & SSIM$\uparrow$ & APD$\downarrow$ & LPIPS$\downarrow$ \\ \midrule
		\multirow{4}{*}{QF=75} & \XSolidBrush & \XSolidBrush & \XSolidBrush & 10.29/12.73 & .229/.134 & 45.80/46.14 & .258/.423 \\
		\multirow{4}{*}{} & \CheckmarkBold & \XSolidBrush & \XSolidBrush & 14.75/11.82 & .584/.274 & 31.87/60.88 & .126/.444 \\
		\multirow{4}{*}{} & \CheckmarkBold & \CheckmarkBold & \XSolidBrush & 24.02/25.51 & .719/.749 & 7.02/10.62 & .053/.109 \\
		\multirow{4}{*}{} & \CheckmarkBold & \CheckmarkBold & \CheckmarkBold & \textbf{31.92}/\textbf{35.38} & \textbf{.923}/\textbf{.935} & \textbf{4.94}/\textbf{3.78} & \textbf{.027}/\textbf{.018} \\ \hline \hline
		\multirow{4}{*}{QF=95} & \XSolidBrush & \XSolidBrush & \XSolidBrush & 10.43/16.77 & .224/.276 & 44.36/26.39 & .263/.292 \\
		\multirow{4}{*}{} & \CheckmarkBold & \XSolidBrush & \XSolidBrush & 12.53/11.75 & .305/.234 & 38.60/61.14 & .222/.423 \\
		\multirow{4}{*}{} & \CheckmarkBold & \CheckmarkBold & \XSolidBrush & 17.48/14.01 & .452/.209 & 33.3/49.93 & .164/.400 \\
		\multirow{4}{*}{} & \CheckmarkBold & \CheckmarkBold & \CheckmarkBold & \textbf{34.03}/\textbf{32.16} & \textbf{.920}/\textbf{.891} & \textbf{4.08}/\textbf{5.21} & \textbf{.013}/\textbf{.046} \\
		\bottomrule
	\end{tabular}
\end{table*}
\subsection{Ablation Study}
The ablation experiments are performed on the ImageNet dataset. Here, we mainly discuss the number of transformer-based invertible sub-module and the effectiveness of the fine-grained DCT representations module and sub-band features enhancement module that greatly impact the final results. As reported in Table \ref{table:SS}, we can observe that the hiding and recovery performance is optimal when the number of transformer-based invertible sub-module is 12. Thus, in our approach, the number of the invertible sub-module $N=12$. Based on this, we further verify the performance effects of the fine-grained DCT representations module and sub-band features enhancement module, as shown in Table \ref{table:AC}. It can be observed that instead of spatial domain data or ordinary DCT data, using fine-grained DCT data as input (the 4-th and 8-th rows) can significantly boost the hiding and revealing performance, and the addition of the sub-band features enhancement module (the 5-th and 9-th rows) can bring further performance improvement.

\section{Conclusions}
In this work, we present a novel and effective model named \textsf{EFDR} for hiding image-level messages within JPEG images, which can fully mine the fine-grained DCT representations being beneficial for hiding and recovery. Besides, we devise to hide and reveal the secret image in the quantized DCT coefficients of cover and stego JPEG images, avoiding the lossy process. To our best knowledge, this is the first attempt to embed a color image of equal size in a color JPEG image. Comprehensive experiments evaluate the efficacy of \textsf{EFDR} and its superior performance both quantitatively and qualitatively. The design of \textsf{EFDR} can provide some inspiration for further research on hiding image-level messages within JPEG covers and extending secret image types to JPEG images.

\clearpage
\bibliographystyle{ACM-Reference-Format}
\bibliography{References}{\balance}

\end{document}